\documentclass[twocolumn,english,leqno]{article}
\usepackage{geometry}
\usepackage{babel}
\usepackage{array}
\usepackage{multirow}
\usepackage{amsmath}
\usepackage{graphicx}

\makeatletter

\providecommand{\tabularnewline}{\\}

\usepackage{amsmath,amstext,amsfonts,amsxtra,amssymb,latexsym,bm}
\usepackage{microtype}
\usepackage{graphicx}
\usepackage{flushend}

\date{}

\makeatother

\begin{document}

\title{Multilevel Anomaly Detection for Mixed Data}

\author{Kien Do\thanks{Centre for Pattern Recognition and Data Analytics, Deakin University,
Geelong, Australia.}~\thanks{dkdo@deakin.edu.au} \and Truyen Tran\thanks{truyen.tran@deakin.edu.au}
\and Svetha Venkatesh\thanks{svetha.venkatesh@deakin.edu.au} }

\maketitle
\global\long\def\model{\mathtt{MIXMAD}}
\global\long\def\xb{\boldsymbol{x}}
\global\long\def\yb{\boldsymbol{y}}
\global\long\def\eb{\boldsymbol{e}}
\global\long\def\zb{\boldsymbol{z}}
\global\long\def\hb{\boldsymbol{h}}
\global\long\def\ab{\boldsymbol{a}}
\global\long\def\bb{\boldsymbol{b}}
\global\long\def\cb{\boldsymbol{c}}
\global\long\def\sigmab{\boldsymbol{\sigma}}
\global\long\def\gammab{\boldsymbol{\gamma}}
\global\long\def\alphab{\boldsymbol{\alpha}}
\global\long\def\rb{\boldsymbol{r}}
\global\long\def\fb{\boldsymbol{f}}
\global\long\def\ib{\boldsymbol{i}}
\global\long\def\Wb{\boldsymbol{W}}
\global\long\def\pb{\boldsymbol{p}}

\begin{abstract}
Anomalies are those deviating from the norm. Unsupervised anomaly
detection often translates to identifying low density regions. Major
problems arise when data is high-dimensional and mixed of discrete
and continuous attributes. We propose $\model$, which stands for
MIXed data Multilevel Anomaly Detection, an ensemble method that estimates
the sparse regions across multiple levels of abstraction of mixed
data. The hypothesis is for domains where multiple data abstractions
exist, a data point may be anomalous with respect to the raw representation
or more abstract representations. To this end, our method sequentially
constructs an ensemble of Deep Belief Nets (DBNs) with varying depths.
Each DBN is an energy-based detector at a predefined abstraction level.
At the bottom level of each DBN, there is a Mixed-variate Restricted
Boltzmann Machine that models the density of mixed data. Predictions
across the ensemble are finally combined via rank aggregation. The
proposed $\model$ is evaluated on high-dimensional real-world datasets
of different characteristics. The results demonstrate that for anomaly
detection, (a) multilevel abstraction of high-dimensional and mixed
data is a sensible strategy, and (b) empirically, $\model$ is superior
to popular unsupervised detection methods for both homogeneous and
mixed data. 
\end{abstract}

\section{Introduction}

A vital intelligent function and survival skill for living organism
is detecting anomalies, that is, those deviating from the norm. Except
for a few deadly instances, we learn to detect anomalies by observing
and exploring, without supervision. Unsupervised anomaly detection
does not assume any domain knowledge about abnormality, and hence
it is cheap and pervasive. A disciplined approach is to identify instances
lying in low density regions \cite{chandola2009anomaly}. However,
estimating density in high-dimensional and mixed-type settings is
difficult \cite{do2016outlier,lud2016iscovering,zimek2012survey}.

Under these conditions, existing non-parametric methods that define
a data cube to estimate the relative frequency of data are likely
to fail. It is because the number of cube grows exponentially with
the data dimensions, thus a cube with only a few or no observed data
points needs not be a low density region. An alternative is to use
distance to $k$ nearest neighbors, assuming that the larger the distance,
the less dense the region \cite{angiulli2002fast}. But distance is
neither well-defined under mixed types nor meaningful in a high-dimensional
space \cite{aggarwal2001surprising,zimek2012survey}. Solving both
challenges is largely missing in the literature as existing work targets
either single-type high-dimensions (e.g., see \cite{zimek2012survey}
for a recent review) or mixed data (e.g., see \cite{do2016outlier,lud2016iscovering}
for latest attempts).

To tackle the challenges jointly, we advocate \emph{learning data
representation through abstraction}, a strategy that (a) transforms
mixed-data into a homogeneous representation \cite{Truyen:2011b},
and (b) represents the multilevel structure of data \cite{bengio2013representation}.
The hypothesis is \emph{a data point may be anomalous with respect
to either the raw representation} \emph{or higher abstractions}. We
call it the \emph{Multilevel Anomaly Detection (MAD)} hypothesis.
For example, an image may be anomalous not because its pixel distribution
differs from the rest, but because its embedded concepts deviate significantly
from the norm. Another benefit of learning higher-level data representation
is that through abstraction, regularities and consistencies may be
readily revealed, making it easier to detect deviations. To test the
MAD hypothesis, we present a new density-based method known as $\model$,
which stands for MIXed data Multilevel Anomaly Detection. $\model$
generalizes the recent work in \cite{do2016outlier} for mixed data
by building multiple abstractions of data. For data abstraction, we
leverage recent advances in unsupervised deep learning to abstract
the data into multilevel low-dimensional representations \cite{bengio2013representation}.

While deep learning has revolutionized supervised prediction \cite{lecun2015deep},
its application to unsupervised anomaly detection is very limited
\cite{zhai2016deep}. With $\model$ we build a sequence of Deep Belief
Nets (DBNs) \cite{hinton2006rdd} of increasing depths. Each DBN is
a layered model that allows multiple levels of data abstraction. The
top layer of the DBN is used as an anomaly detector. A key observation
to DBN-based anomaly detection is that the perfect density estimation
is often not needed in practice. All we need is a ranking of data
instances by increasing order of abnormality. An appropriate anomaly
scoring function is \emph{free-energy of abstracted data}, which equals
negative-log of data density up to an additive constant.

$\model$ offers the following procedure to test the MAD hypothesis:
First apply multiple layered abstractions to the data, and then estimate
the anomalies at each level. Finally, the anomaly score is aggregated
across levels. While this bears some similarity with the recent ensemble
approaches \cite{aggarwal2015theoretical,ando2015ensemble}, the key
difference is MAD relies on multiple data abstractions, not data resampling
or random subspaces which are still on the original data level. In
$\model$, as the depth increases and the data representation is more
abstract, the energy landscape gets smoother, and thus it may detect
different anomalies. For reaching anomaly consensus across depth-varying
DBNs, $\model$ employs a simple yet flexible rank aggregation method
based on $p$-norm. 

We validate $\model$ through an extensive set of experiments against
well-known shallow baselines, which include the classic methods (PCA,
GMM, RBM and one-class SVM), as well as state-of-the-art mixed-type
methods (ODMAD \cite{koufakou2008detecting}, BMM \cite{bouguessa2015practical},
GLM-t \cite{lud2016iscovering} and Mv.RBM \cite{do2016outlier}).
The experiments demonstrate that (a) multilevel abstraction of data
is important in anomaly detection, and (b) $\model$ is superior to
popular unsupervised detection methods for both homogeneous and mixed
data. 

In summary, we claim the following contributions:
\begin{itemize}
\item Stating the hypothesis of Multilevel Anomaly Detection (MAD) that
argues for reaching agreement across multiple abstractions of data.
\item Deriving $\model$, an efficient ensemble algorithm to test MAD. $\model$
builds a sequence of Deep Belief Nets, each of which is an anomaly
detector. All detectors are then combined using a flexible $p$-norm
aggregation that allows tuning along the conservative/optimistic axis.
\item A comprehensive evaluation of $\model$ on high-dimensional datasets
against a large suite of competing methods. 
\end{itemize}

\section{Background \label{sec:Background}}

Anomaly detection on high-dimensional and mixed data has attracted
a wide range of methods, but the two challenges are tackled independently
rather than jointly as in this paper. High-dimensional data suffers
from `curse of dimensionality' also known more concretely as `distance
concentration effect', irrelevant attributes and redundant attributes,
which together cause failure of low-dimensional techniques \cite{aggarwal2001surprising}.
Popular anomaly detection approaches targeting high-dimensions include
feature selection, dimensionality reduction (such as using PCA) and
subspace analysis (readers are referred to \cite{zimek2012survey}
for a recent survey and in-depth discussion).

Mixed data has received a moderate attention. A method called LOADED
\cite{ghoting2004loaded} defines the score on the discrete subspace
and combines with a correlation matrix in the continuous subspace.
A related method called ODMAD \cite{koufakou2008detecting} opts for
stage-wise detection in each subspace. A different strategy is employed
in \cite{bouguessa2015practical}, where scores for discrete and continuous
spaces are computed separately, then combined using a mixture model.
The work in \cite{zhang2010effective} introduces Pattern-based Anomaly
Detection (POD), where a pattern consists of a discrete attribute
and all continuous attributes. Scores of all patterns are then combined.
Methods with joint distribution of all attributes are introduced recently
in \cite{do2016outlier,lud2016iscovering} using latent variables
to link all types together. We adapt the work in \cite{do2016outlier}
to represent mixed data into a homogeneous form using Mixed-variate
Restricted Boltzmann Machines \cite{Truyen:2011b}. The homogeneous
representation can then be abstracted into higher semantic levels
in order to test the MAD hypothesis stated earlier. 

The recent advances of deep networks have inspired some work in anomaly
detection \cite{becker2015deep,gao2014intrusion,sun2014automated,tagawa2014structured,wang2016deep}.
A common strategy is to use unsupervised deep networks to detect features,
which are then fed into well-established detection algorithms \cite{wang2016deep}.
Another strategy is to learn a deep autoencoder which maps data to
itself, and then use the reconstruction error as anomaly score \cite{becker2015deep,gao2014intrusion,sun2014automated,tagawa2014structured}.
A problem with this reconstruction error is that the final model still
operates on raw data, which can be noisy and high-dimensional. A more
fundamental problem is that reconstruction error does not reflect
data density \cite{kamyshanska2015potential}. A better approach is
to use deep networks to estimate the energy directly \cite{do2016outlier,fiore2013network,zhai2016deep}.
This resembles in principle our free-energy function presented in
Sec\@.~\ref{subsec:Abstracted-Anomaly-Detection}, but differs in
the network construction procedure. It does not reflect the multilevel
abstraction hypothesis we are pursuing.

\section{$\protect\model$: MIXed data Multilevel Anomaly Detection \label{sec::-MIXed-data}}

We present $\model$, an ensemble algorithm for \textbf{MIX}ed data\textbf{
M}ultilevel \textbf{A}nomaly \textbf{D}etection. Given a data instance
$\xb$ we estimate the unnormalized density, which is the true density
up to a multiplicative constant: $\tilde{P}(\xb)=cP(\xb)$. An instance
is declared as anomaly if its density is lower than a threshold:
\begin{equation}
\log\tilde{P}(\xb)\le\beta\label{eq:outlier-decision}
\end{equation}
for some threshold $\beta$ estimated from data. Here $-\log\tilde{P}(\xb)$
serves as the anomaly scoring function.

\begin{figure}
\begin{centering}
\includegraphics[width=1\columnwidth]{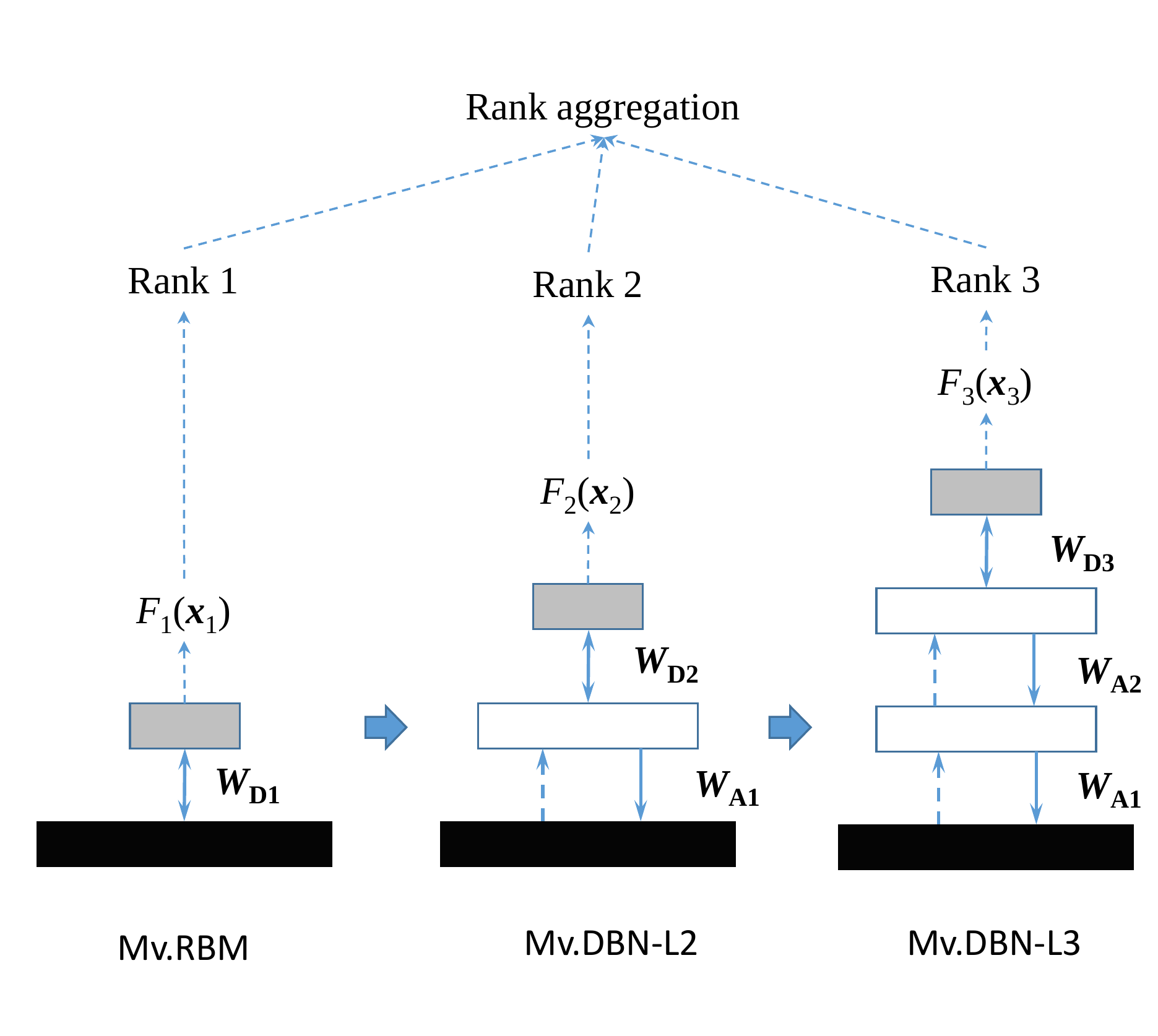}
\par\end{centering}
\caption{Multilevel anomaly detection based on successive DBNs. DBNs are ``grown''
sequentially (left to right), with abstraction layer inserted. Filled
boxes represent data input, empty boxes represent abstraction layers,
and shaded boxes represent the hidden layer of the detection RBM.\label{fig:MAD}}
\end{figure}

\subsection{Prelim: Shallow Model for Mixed Data\@.~}

For subsequent development, let us briefly review a probabilistic
graphical model known as Mixed-variate Restricted Boltzmann Machines
(Mv.RBM) \cite{Truyen:2011b} for modelling mixed data. Let $\xb$
be an input vector of $N$ (mixed-type) elements, and $\hb\in\{0,1\}^{K}$
be a binary hidden vector, Mv.RBM defines the joint distribution as
follows:
\[
P\left(\xb,\hb\right)\propto\exp\left(-E(\xb,\hb)\right)
\]
where $E(\xb,\hb)$ is energy function decomposed as follows:
\begin{equation}
E(\xb,\hb)=\sum_{i=1}^{N}E_{i}(x_{i})+\sum_{k=1}^{K}\left(-b_{k}+\sum_{i=1}^{N}G_{ik}(x_{i})\right)h_{k}\label{eq:Mv.RBM-energy}
\end{equation}
The sub-energy functions $E_{i}(x_{i})$ and $G_{ik}(x_{i})$ are
type-specific. For example, $E_{i}(x_{i})=-a_{i}x_{i}$ for binary;
$E_{i}(x_{i})=0.5x_{i}^{2}-a_{i}x_{i}$ for Gaussian; and $\log x_{i}!-a_{i}x_{i}$
for Poisson. The three types share the same form of input-hidden mapping:
$G_{ik}(x_{i})=W_{ik}x_{i}$. 

Mv.RBM can be used for outlier detection \cite{do2016outlier} by
noticing that:
\begin{equation}
P(\xb)\propto\sum_{\hb}\exp\left(-E(\xb,\hb)\right)=\exp\left(-F(\xb)\right)\label{eq:data-marginal}
\end{equation}
where $F(\xb)=-\log\sum_{\hb}\exp\left(-E(\xb,\hb)\right)$ is known
as \emph{free-energy}. In other words, $F(\xb)=-\log P(\xb)+c$ or
equivalently $F(\xb)=-\log\tilde{P}(\xb)$, where $c$ is a constant
and $\tilde{P}(\xb)$ is unnormalized density (see Eq.~(\ref{eq:outlier-decision})).
Thus \emph{we can use the free-energy as an anomaly score} to rank
data instances, following the decision rule in Eq.~(\ref{eq:outlier-decision}).
The computational cost of free-energy scales linearly with number
of dimensions making it an attractive scoring in practice: 

\begin{equation}
F(\xb)=\sum_{i=1}^{N}E_{i}(x_{i})-\sum_{k=1}^{K}\log\left(1+\exp\left(b_{k}-\sum_{i=1}^{N}G_{ik}(x_{i})\right)\right)\label{eq:free-energy-Mv.RBM}
\end{equation}

For training, we adopt the standard CD-1 procedure \cite{Hinton02}.
This method approximates the gradient of the data log-likelihood using
one random sample per data point. The sample is generated using one-step
MCMC starting from the observed data itself.

\subsection{Extending Deep Belief Nets for Mixed Data.\label{subsec:Deep-Belief-Nets}~}

Deep Belief Network (DBN) \cite{hinton2006rdd} is a generative model
of data. It assumes that the data $\xb$ is generated from hidden
binary variables $\hb^{1}$, which is generated from higher hidden
binary variables $\hb^{2}$ and so on. Two consecutive layers in DBN
form a Restricted Boltzmann Machine (RBM), which models either $P\left(\xb,\hb^{1}\right)$
at the bottom level, or $P\left(\hb^{l},\hb^{l+2}\right)$ at higher
levels. A DBN is usually trained by learning a stack of RBMs in a
layer-wise fashion. First a RBM is trained on the input data, its
weights are then frozen. The hidden posterior is used to generate
input for the next RBM, i.e., $\hb\sim P\left(\hb^{1}\mid\xb\right)$.
The process is repeated until the last RBM. This procedure of freezing
the lower weights has been shown to optimize the variational bound
of the data likelihood $\log P(\xb)$ \cite{hinton2006fast}. Overall,
an DBN is a mixed-graph whose the top RBM remains undirected, but
the lower cross-layer connections are directed toward the data.

The original DBNs are designed for single data type, primarily binary
or Gaussian. Here we extend DBNs to accommodate mixed data. In particular,
the training steps of DBNs are kept, but the bottom RBM is now a Mv.RBM.
The Mv.RBM transforms mixed input $\xb$ into a homogeneous binary
representation through $\hb\sim P\left(\hb^{1}\mid\xb\right)$. The
subsequent RBMs are for binary inputs as usual.

\subsection{Abstracted Anomaly Detection Using Deep Belief Nets.~ \label{subsec:Abstracted-Anomaly-Detection}}

Although the stagewise learning procedure that gives rise to DBN optimizes
the lower bound of $\log P(\xb)$, it is still not possible to estimate
the bound for density-based anomaly detection. Let $L$ be the number
of hidden layers. Existing methods typically use DBNs to (a) learn
high-level features through $P\left(\hb_{L}\mid\hb_{L-1}\right)$
and feed to existing anomaly detectors (e.g., \cite{wang2016deep});
and (b) build a deep autoencoder then estimate the reconstruction
error \cite{becker2015deep,gao2014intrusion,sun2014automated,tagawa2014structured}.
Here we propose an alternative to use DBN \emph{directly} for anomaly
detection.

The idea is to recognize that the RBM at the top of the DBN operates
on data abstraction $\hb_{L}$, and the RBM's prior density $P_{L}\left(\hb_{L}\right)$
can replace $P(\xb)$ in Eq.~(\ref{eq:data-marginal}). Recall that
the input $\hb_{l}$ to the intermediate RBM at level $l$ is an abstraction
of the lower level data. It is generated through sampling the posterior
$P\left(\hb_{l}\mid\hb_{l-1}\right)$ for $l\ge2$ and $P\left(\hb_{l}\mid\xb\right)$
for $l=1$ as follows: 
\begin{equation}
\hb_{l}\sim\begin{cases}
\mathcal{B}\left(\sigma\left(\bb_{l-1}+\Wb_{l-1}\hb_{l-1}\right)\right) & \text{for}\,l\ge2\\
\mathcal{B}\left(\sigma\left(b_{1k}-\sum_{i=1}^{N}G_{ik}(x_{i})\right)\right) & \text{for}\,l=1
\end{cases}\label{eq:abstraction}
\end{equation}
where $\text{\ensuremath{\mathcal{B}}}$ stands for Bernoulli distribution
and $\sigma(z)=\left[1+e^{-z}\right]^{-1}$. The prior density $P_{L}\left(\hb_{L}\right)$
can be rewritten as:
\begin{equation}
P\left(\hb_{L}\right)\propto\exp\left(-F_{L}(\hb_{L})\right)\label{eq:log-P}
\end{equation}
where
\begin{equation}
F_{L}\left(\hb_{L}\right)=-\bb_{L}^{'}\hb_{L}-\sum_{k=1}^{K_{L}}\log\left(1+\exp\left(\ab_{Lk}+\Wb_{Lk}\hb_{L}\right)\right)\label{eq:free-energy-MAD}
\end{equation}
This abstracted\emph{ }free-energy, like that of Mv.RBM in Eq.~(\ref{eq:free-energy-Mv.RBM}),
can also be used as an anomaly score of abstracted data, and the anomaly
region is defined as:
\[
\mathcal{R}\in\left\{ \xb\mid F_{L}\left(\hb_{L}\right)\ge\beta\right\} 
\]

Once the DBN has been trained, the free-energy can be approximated
by a deterministic function, where the intermediate input $\hb_{l}$
in Eq.~(\ref{eq:abstraction}) is recursively replaced by: 
\begin{equation}
\hb_{l}\approx\begin{cases}
\sigma\left(\bb_{l-1}+\Wb_{l-1}\hb_{l-1}\right) & \,\,\text{for}\,l\ge2\\
\sigma\left(b_{1k}-\sum_{i=1}^{N}G_{ik}(x_{i})\right) & \,\,\text{for}\,l=1
\end{cases}\label{eq:recursion}
\end{equation}

\subsection{Multilevel Detection Procedure With DBN Ensemble.~}

Recall that our Multilevel Anomaly Detection (MAD) hypothesis is that
for domains where multiple data abstractions exist (e.g., in images
\& videos), an anomaly can be detected on one or more abstract representations.
Each level of abstraction would detect abnormality in a different
way. For example, assume an indoor setting where normal images contain
regular arrangement of furniture. An image of a room with random arrangement
(e.g., a chair in a bed) may appear normal at the pixel level, and
at the object class level, but not at the object context level. This
suggests the following procedure: \emph{apply multiple abstraction
levels, and at each level, estimate an anomaly score, then combine
all the scores}.

Since free-energies in Eq.~(\ref{eq:free-energy-MAD}) differ across
levels, direct combination of anomaly scores is not possible. A sensible
approach is through rank aggregation, that is, the free-energies at
each level are first used to rank instances from the lowest to the
highest energy. The ranks now serve as anomaly scores which are compatible
across levels.

\subsubsection{$p$-norm Rank Aggregation.~}

One approach to rank aggregation is to find a ranking that minimizes
the disagreement with all ranks \cite{ailon2008aggregating}. However,
this minimization requires searching through a permutation space of
size $N!$ for $N$ instances, which is intractable. Here we resort
to a simple technique: Denoted by $r_{li}\ge0$ the rank anomaly score
of instance $i$ at level $l$, the aggregation score is computed
as:
\begin{equation}
\bar{r}_{i}(p)=\left(\sum_{l=1}^{L}r_{li}^{p}\right)^{1/p}\label{eq:p-norm}
\end{equation}
where $p>0$ is a tuning parameter.

There are two detection regimes under this aggregation scheme. The
detection at $p<1$ is \emph{conservative}, that is, individual high
outlier scores are suppressed in favor of a consensus. The other regime
is \emph{optimistic} at $p>1$, where the top anomaly scores tend
to dominate the aggregation. This aggregation subsumes several methods
as special cases: $p=1$ reduces to the classic Borda count when $s_{li}$
is rank position; $p=\infty$ reduces to the max: $\lim_{p\rightarrow\infty}\bar{r}_{i}(p)=\max_{l}\left\{ r_{li}\right\} $. 

\subsubsection{Separation of Abstraction and Detection.~\label{subsec:Separation-of-Abstraction}}

Recall from that we use RBMs for both abstraction (Eq.~(\ref{eq:abstraction}))
and anomaly detection (Eq.~(\ref{eq:free-energy-MAD})). Note that
data abstraction and anomaly detection have different goals \textendash{}
abstraction typically requires more bits to adequately disentangle
multiple factors of variation \cite{bengio2013representation}, whereas
detection may require less bits to estimate a rank score. Fig.~\ref{fig:algorithm}
presents the multilevel anomaly detection algorithm. It trains one
Mv.RBM and $(L-1)$ DBNs of increasing depths \textendash{} from $2$
to $L$ \textendash{} with time complexity linear in $L$. They produce
$L$ rank lists, which are then aggregated using Eq.~(\ref{eq:p-norm}).

\begin{figure}
\begin{centering}
\begin{tabular}{|>{\raggedright}p{0.95\columnwidth}|}
\hline 
\textbf{Input}: data $\mathcal{D}=\left\{ \xb\right\} $; \textbf{Output}:
Anomaly rank.\\

\textbf{User-defined parameters}: depth $L$, abstraction hidden sizes
$\left\{ K_{1}^{a},K_{2}^{a},..K_{L-1}^{a}\right\} $, and detection
hidden sizes $\left\{ K_{1}^{d},K_{2}^{d},..K_{L}^{d}\right\} $,
and $p$.\\

\textbf{Main-loop:}
\begin{enumerate}
\item For each level $l=1,2,...,L$:

\begin{enumerate}
\item Train a \emph{detection RBM} (or \emph{Mv.RBM} if $l=1$) with $K_{l}^{d}$
hidden units;
\item Estimate free-energy $F_{l}(\hb_{l})$ using Eqs.~(\ref{eq:free-energy-MAD},\ref{eq:recursion})
(or $F(\xb)$ using Eq.~(\ref{eq:free-energy-Mv.RBM}) for $l=1$);
\item Rank data according to free energies;
\item If $l<L$

\begin{enumerate}
\item Train an \emph{abstraction RBM} on $\hb_{l}$ (or $\xb$ for if $l=1$)
with $K_{l}^{a}$ hidden units;
\item Abstracting data using Eq.~(\ref{eq:abstraction}) to generate $\hb_{l+1}$;
\end{enumerate}
\end{enumerate}
\item Aggregate ranks using $p$-norm in Eq.~(\ref{eq:p-norm}).
\end{enumerate}
\tabularnewline
\hline 
\end{tabular}
\par\end{centering}
\caption{Multilevel anomaly detection algorithm.\label{fig:algorithm}}
\end{figure}

\section{Experiments \label{sec:Experiments}}

This section reports experiments and results of $\model$ on a comprehensive
suite of datasets. We first present the cases for single data type
in Section~\ref{subsec:Homogeneous-Data}, then extend for mixed
data in Section~\ref{subsec:Mixed-Data}.

\subsection{Homogeneous Data\@.~ \label{subsec:Homogeneous-Data}}

We use three \emph{high-dimensional} real-world datasets with very
different characteristics: \emph{handwritten digits} \emph{(MNIST)},
\emph{Internet ads} and \emph{clinical records of birth episodes.}
\begin{itemize}
\item The \emph{MNIST} has $60,000$ gray images of size $28\times28$ for
training and $10,000$ images for testing\footnote{http://yann.lecun.com/exdb/mnist/}.
The raw pixels are used as features (784 dimensions). Due to ease
of visualization and complex data topology, this is an excellent data
for testing anomaly detection algorithms. We use digit '8' as normal
and a small portion (\textasciitilde{}5\%) of other digits as outliers.
This proves to be a challenging digit compared to other digits \textendash{}
see Fig.~\ref{fig:Detection-on-MNIST} (left) for failure of pixel-based
$k$-nearest neighbor. We randomly pick 3,000 training images and
keep all the test set.
\item The second dataset is \emph{InternetAds} with 5\% anomaly injection
as described in \cite{campos2015evaluation}. As the data size is
moderate (1,682 instances, 174 features), no train/test splitting
is used. 
\item The third dataset consists of \emph{birth episodes} collected from
an urban hospital in Sydney, Australia in the period of 2011\textendash 2015
\cite{tran2016preterm}. Preterm births are considered anomalous as
they have a serious impact on the survival and development of the
babies \cite{vovsha2014predicting}. In general, births occurring
within 37 weeks of gestation are considered preterm \cite{barros2015distribution}.
We are also interested in early preterm births, e.g., those occurring
within 34 weeks of gestation. This is because the earlier the birth,
the more severe the case, leading to more intensive care. Features
include 369 clinically relevant facts collected in the first few visits
to hospital before 25 weeks of gestation. The data is randomly split
into a training set of 3,000 cases, and a test set of 5,104 cases.
\end{itemize}
All data are normalized into the range {[}0,1{]}, which is known to
work best in \cite{campos2015evaluation}. Data statistics are reported
in Table~\ref{tab:Data-statistics}.

\begin{table}
\begin{centering}
\begin{tabular}{|l|c|c|c|c|}
\hline 
 & Dims & \#train & \#test & \%outlier\tabularnewline
\hline 
\hline 
\emph{MNIST} & 784 & 3,000 & 1,023 & 4.9\tabularnewline
\hline 
\emph{InternetAds} & 174 & 1,682 & 1,682 & 5.0\tabularnewline
\hline 
\emph{Preterm-37w} & 369 & 3,000 & 5,104 & 10.9\tabularnewline
\hline 
\emph{Preterm-34w} & 369 & 3,000 & 5,104 & 6.5\tabularnewline
\hline 
\end{tabular}
\par\end{centering}
\caption{Homogeneous data statistics.\label{tab:Data-statistics}}
\end{table}

\subsubsection{Models implementation.~}

We compare the proposed method against four popular shallow unsupervised
baselines \textendash{} $k$-\emph{NN}, \emph{PCA}, \emph{Gaussian
mixture model (GMM)}, and \emph{one-class SVM (OCSVM)} \cite{chandola2009anomaly}.
(a) The $k$-NN uses the mean distance from a test case to the $k$
nearest instances as outlier score \cite{angiulli2002fast}. We set
$k=10$ with Euclidean distance. (b) For PCA, the $\alpha$\% total
energy is discarded, where $\alpha$ is the estimated outlier rate
in training data. The reconstruction error using the remaining eigenvectors
is used as the outlier score. (c) The GMMs have four clusters and
are regularized to work with high dimensional data. The negative log-likelihood
serves as outlier score. (d) The OCVSMs have RBF kernels with automatic
scaling. We also consider RBM \cite{fiore2013network} as baseline,
which is a special case of our method where the number of layers is
set to $L=1$. 

\begin{table}
\begin{centering}
\begin{tabular}{|c|c|c|c|}
\hline 
Param. & \emph{MNIST} & \emph{InternetAds} & \emph{Preterm}\tabularnewline
\hline 
\hline 
$K_{D}$ & 10 & 10 & 10\tabularnewline
\hline 
$K_{A}$ & 70 & 50 & 70\tabularnewline
\hline 
$N$ & 784 & 174 & 369\tabularnewline
\hline 
\end{tabular}
\par\end{centering}
\caption{Settings of the $\protect\model$ for homogeneous data. $K_{D}$ and
$K_{A}$ are the number of hidden units in the detection RBM and the
abstraction RBMs respectively, and $N$ is data dimension.\label{tab:Settings-of-MAD}. }
\end{table}

For $\model$, abstraction RBMs have the same number of hidden units
while detection RBM usually have smaller number of hidden units. All
RBMs are trained using CD-1 \cite{Hinton02} with batch size of 64,
learning rate of 0.3 and 50 epochs. Table~\ref{tab:Settings-of-MAD}
lists model parameters used in experimentation.

We use the following evaluation measures: \emph{Area Under ROC Curve
(AUC)}, and \emph{NDCG@T}. The AUC reflects the average discrimination
power across the entire dataset, while the NDCG@T places more emphasis
on the top retrieved cases.

\subsubsection{Results.~}

\begin{figure}
\begin{centering}
\begin{tabular}{|c|c|c|}
\cline{1-1} \cline{3-3} 
\includegraphics[width=0.42\columnwidth,height=0.4\columnwidth]{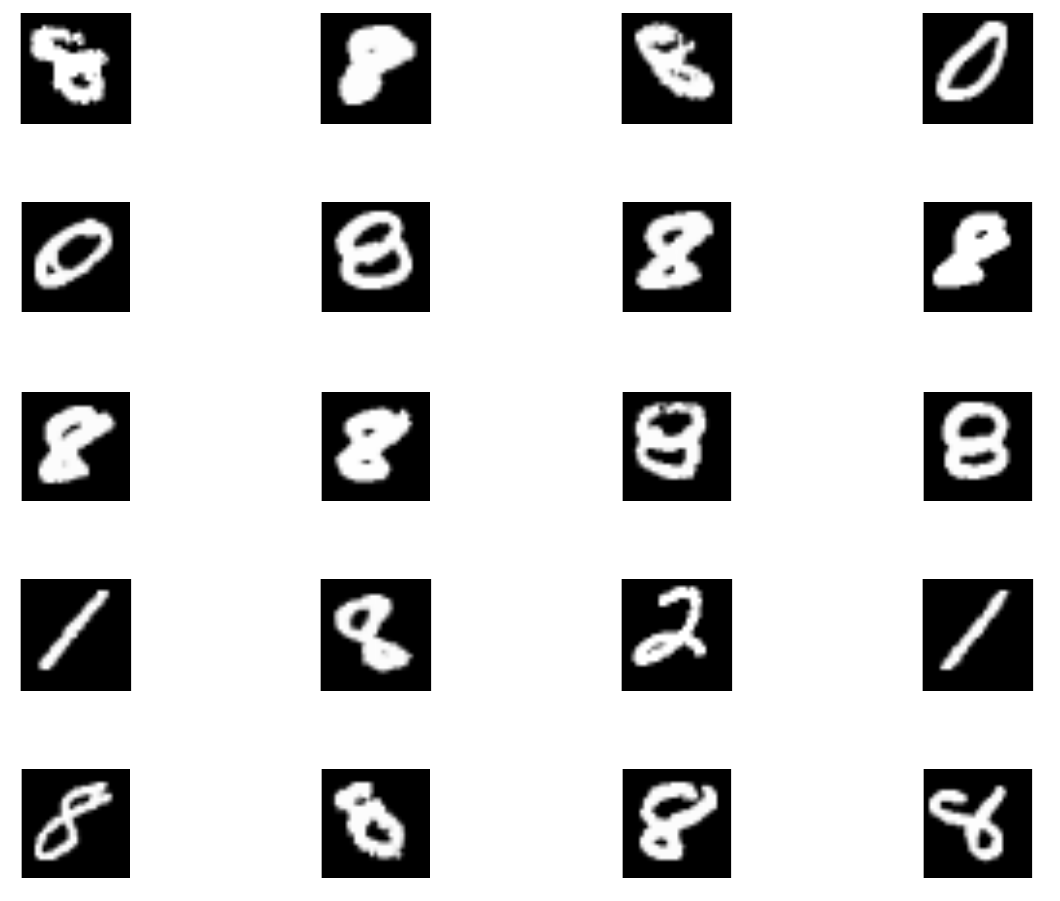} &  & \includegraphics[width=0.42\columnwidth,height=0.4\columnwidth]{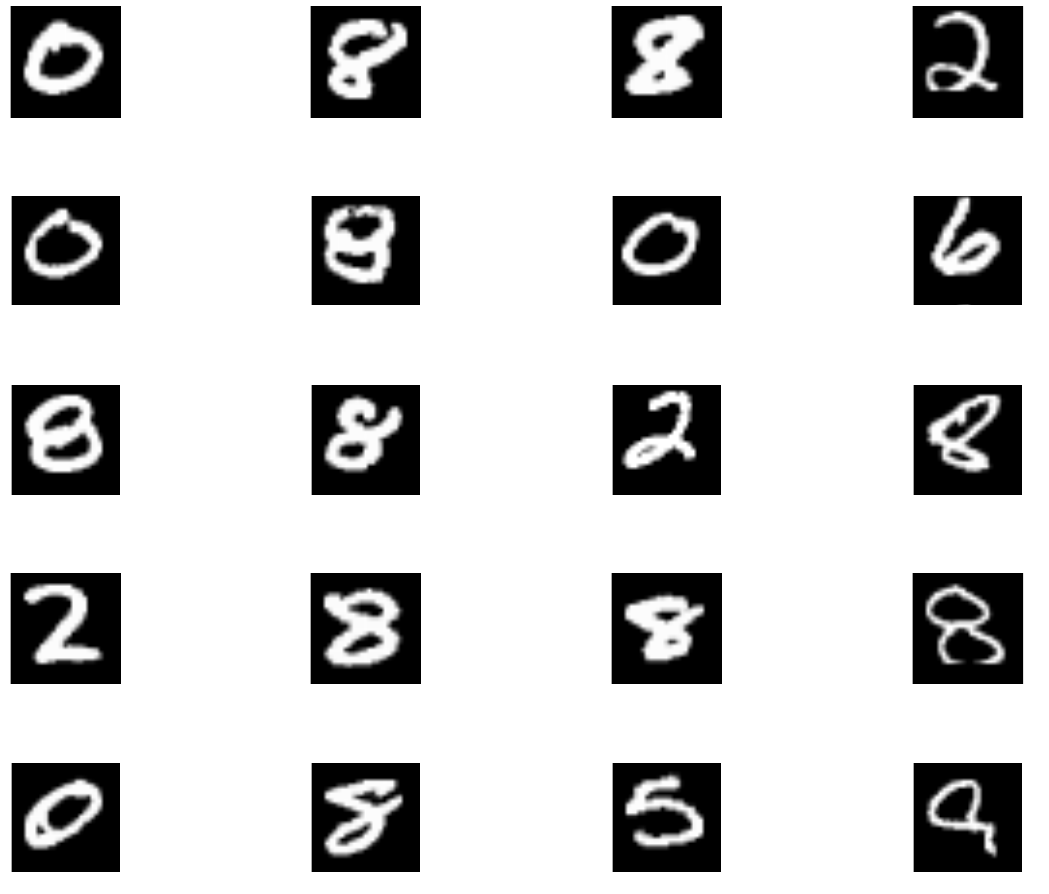}\tabularnewline
$k$-NN, err: 15 &  & RBM, err: 10\tabularnewline
\cline{1-1} \cline{3-3} 
\multicolumn{1}{c}{} & \multicolumn{1}{c}{} & \multicolumn{1}{c}{}\tabularnewline
\cline{1-1} \cline{3-3} 
\includegraphics[width=0.42\columnwidth,height=0.4\columnwidth]{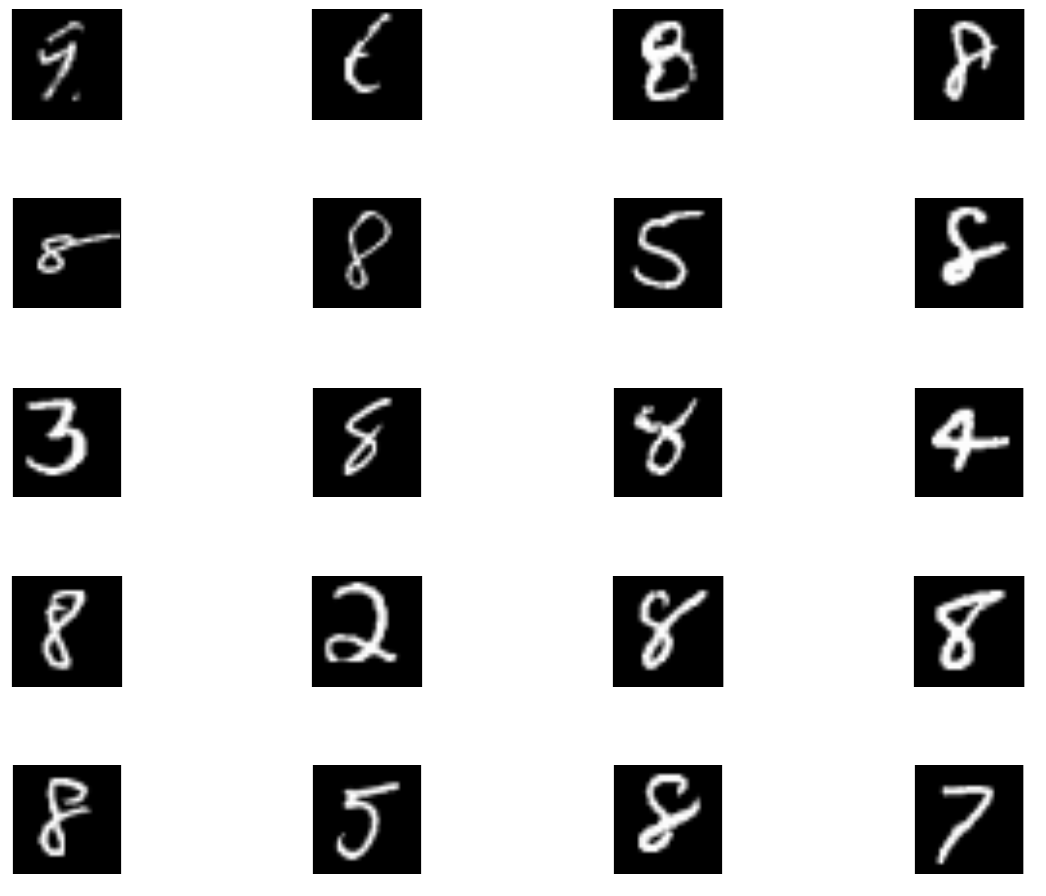} &  & \includegraphics[width=0.42\columnwidth,height=0.4\columnwidth]{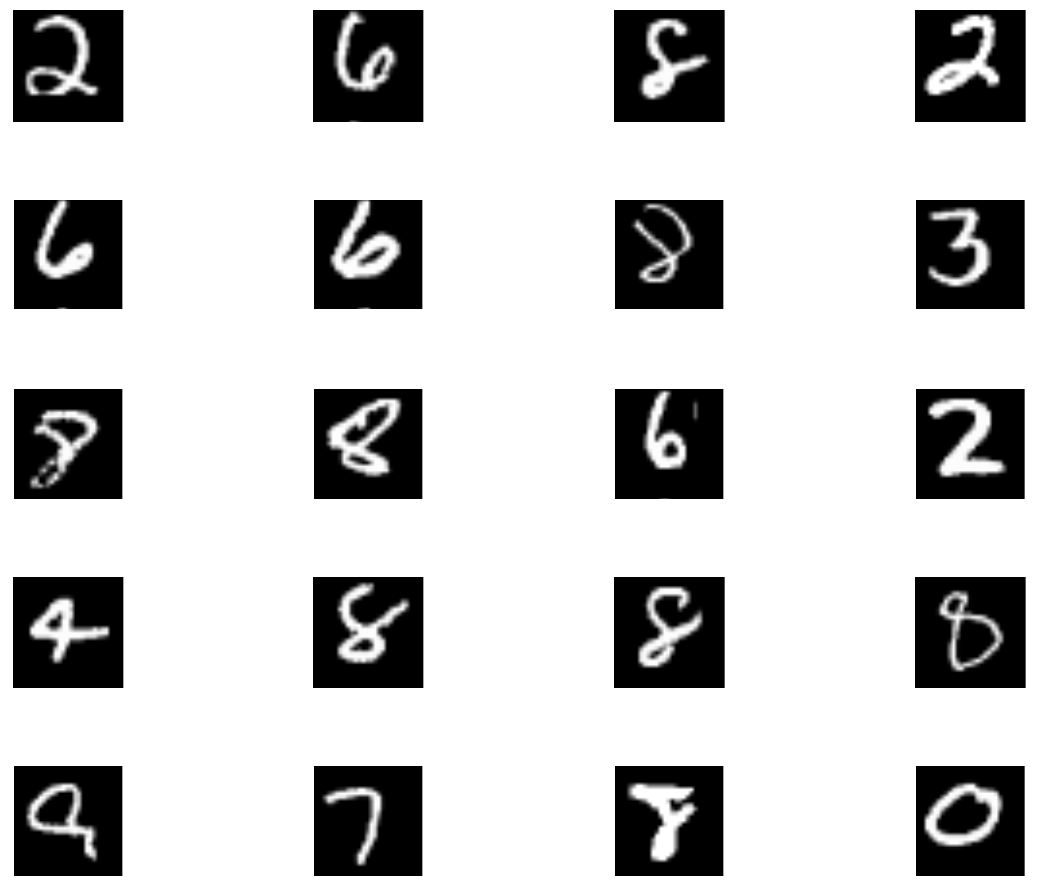}\tabularnewline
DBN-L2, err: 12 &  & $\model$-L2p2, err: 8\tabularnewline
\cline{1-1} \cline{3-3} 
\end{tabular}
\par\end{centering}
\caption{Anomaly detection on MNIST test set for the top 20 digits. Normal
digit is ``8''. L2p2 stands for $L=2$ (two layers) and $p=2$ in
Eq.~(\ref{eq:p-norm}).\label{fig:Detection-on-MNIST}}
\end{figure}

To see how $\model$ combines evidences from detectors in the ensemble,
we run the algorithms: RBM, the DBN with 2 layers, and the $\model$
that combines RBM and DBN results. Fig.~\ref{fig:Detection-on-MNIST}
plots detected images by the RBM/DBN/$\model$ against the classic
$k$-NN. The $k$-NN fails for 15 out of 20 cases, mostly due to the
variation in stroke thickness, which is expected for pixel-based matching.
The RBM and DBN have different errors, confirming that anomalies differ
among abstraction levels. Finally, the ensemble of RBM/DBN, then $\model$
improves the detection significantly. The error is mostly due to the
high variation in styles (e.g., 8 with open loops).

Table~\ref{tab:AUC} reports the Area Under the ROC Curve (AUC) for
all methods and datasets. Overall $\model$ with 2 or 3 hidden layers
works well. The difference between the baselines and $\model$ is
amplified in the NDCG measure, as shown in Table~\ref{tab:NDCG}.
One possible explanation is that the $\model$ is an ensemble \textendash{}
an outlier is considered outlier if it is detected by all detectors
at different abstraction levels. One exception is the max-aggregation
(where $p\rightarrow\infty$ in Eq.~(\ref{eq:p-norm})), where the
detection is over-optimistic.

\begin{table*}
\begin{centering}
\begin{tabular}{|l|c|c|c|c|}
\hline 
\emph{Method} & \emph{MNIST} & \emph{InternetAds} & \emph{Preterm (37wks)} & \emph{Preterm (34wks)}\tabularnewline
\hline 
\hline 
$k$-NN & 0.804 & 0.573 & 0.596 & 0.624\tabularnewline
PCA & 0.809 & 0.664 & 0.641 & 0.673\tabularnewline
GMM & 0.839 & 0.725 & 0.636 & 0.658\tabularnewline
OCSVM & 0.838 & 0.667 & 0.646 & 0.676\tabularnewline
RBM & 0.789 & 0.712 & 0.648 & 0.677\tabularnewline
\hline 
\hline 
$\model$-L2p.5 & \textbf{0.867} & \textbf{0.829} & 0.627 & \textbf{0.729}\tabularnewline
$\model$-L2p1 & \textbf{0.880} & \textbf{0.827} & \textbf{0.645} & \textbf{0.748}\tabularnewline
$\model$-L2p2 & \textbf{0.897} & \textbf{0.816} & \textbf{0.661} & \textbf{0.761}\tabularnewline
$\model$-L2p$\infty$ & \textbf{0.892} & \textbf{0.765} & \textbf{0.660} & \textbf{0.745}\tabularnewline
\hline 
\hline 
$\model$-L3p.5 & 0.787 & \textbf{0.789} & \textbf{0.674} & \textbf{0.757}\tabularnewline
$\model$-L3p1 & 0.814 & \textbf{0.775} & \textbf{0.689} & \textbf{0.765}\tabularnewline
$\model$-L3p2 & \textbf{0.847} & \textbf{0.758} & \textbf{0.685} & \textbf{0.759}\tabularnewline
$\model$-L3p$\infty$ & \textbf{0.876} & \textbf{0.734} & \textbf{0.668} & \textbf{0.742}\tabularnewline
\hline 
\end{tabular}
\par\end{centering}
\caption{The Area Under the ROC Curve (AUC) for homogeneous data. $L$ is the
number of hidden layers, $p$ is the aggregation parameter in Eq.~(\ref{eq:p-norm}),
bold indicate better performance than baselines. Note that RBM is
the limiting case of $\protect\model$ with $L=1$. \label{tab:AUC}}
\end{table*}

\begin{table*}
\begin{centering}
\begin{tabular}{|l|c|c|c|c|}
\hline 
\emph{Method} & \emph{MNIST} & \emph{InternetAds} & \emph{Preterm (37wks)} & \emph{Preterm (34wks)}\tabularnewline
\hline 
\hline 
$k$-NN & 0.218 & 0.413 & 0.362 & 0.188\tabularnewline
PCA & 0.488 & 0.225 & 0.505 & 0.356\tabularnewline
GMM & 0.458 & 0.415 & 0.438 & 0.223\tabularnewline
OCSVM & 0.423 & 0.094 & 0.471 & 0.172\tabularnewline
RBM & 0.498 & 0.421 & 0.429 & 0.216\tabularnewline
\hline 
\hline 
MAD-L2p.5 & \textbf{0.666} & \textbf{0.859} & \textbf{0.945} & \textbf{0.831}\tabularnewline
MAD-L2p1 & \textbf{0.667} & \textbf{0.859} & \textbf{0.945} & \textbf{0.831}\tabularnewline
MAD-L2p2 & \textbf{0.666} & \textbf{0.859} & \textbf{0.945} & \textbf{0.831}\tabularnewline
MAD-L2p$\infty$ & \textbf{0.536} & 0.271 & \textbf{0.741} & \textbf{0.576}\tabularnewline
\hline 
\hline 
MAD-L3p.5 & \textbf{0.732} & \textbf{0.908} & \textbf{0.798} & \textbf{0.625}\tabularnewline
MAD-L3p1 & \textbf{0.732} & \textbf{0.908} & \textbf{0.798} & \textbf{0.626}\tabularnewline
MAD-L3p2 & \textbf{0.732} & \textbf{0.902} & \textbf{0.769} & \textbf{0.597}\tabularnewline
MAD-L3p$\infty$ & 0.360 & \textbf{0.598} & 0.370 & 0.113\tabularnewline
\hline 
\end{tabular}
\par\end{centering}
\caption{The NDCG@20 for homogeneous data. $L$ is the number of hidden layers,
$p$ is the aggregation parameter in Eq.~(\ref{eq:p-norm}), bold
indicate better performance than baselines. Note that RBM is the limiting
case of $\protect\model$ with $L=1$. \label{tab:NDCG}}
\end{table*}

\subsection{Mixed Data\@.~ \label{subsec:Mixed-Data}}

\begin{table*}
\begin{centering}
\begin{tabular}{|l|r|r|c|c|c|c|c|}
\hline 
\multirow{2}{*}{Dataset} & \multicolumn{2}{c|}{No. Instances} & \multicolumn{5}{c|}{No. Attributes}\tabularnewline
\cline{2-8} 
 & \multicolumn{1}{c|}{Train} & Test & Bin. & Gauss. & Nominal & Poisson & Total\tabularnewline
\hline 
\hline 
\emph{KDD99-10} & 75,669 & 32,417 & 4 & 15 & 3 & 19 & 41\tabularnewline
\hline 
\emph{Australian Credit} & 533 & 266 & 3 & 6 & 5 & 0 & 14\tabularnewline
\hline 
\emph{German Credit} & 770 & 330 & 2 & 7 & 11 & 0 & 20\tabularnewline
\hline 
\emph{Heart} & 208 & 89 & 3 & 6 & 4 & 0 & 13\tabularnewline
\hline 
\emph{Thoracic Surgery} & 362 & 155 & 10 & 3 & 3 & 0 & 16\tabularnewline
\hline 
\emph{Auto MPG} & 303 & 128 & 0 & 5 & 3 & 0 & 8\tabularnewline
\hline 
\emph{Contraceptive} & 1136 & 484 & 3 & 0 & 4 & 1 & 8\tabularnewline
\hline 
\end{tabular}
\par\end{centering}
\caption{\label{tab:dataset-prop}Statistics of mixed data. The proportion
of outliers are 10\%.}
\end{table*}

\begin{table*}
\begin{centering}
\begin{tabular}{|l|c|c|c|c|c|c|c|}
\hline 
 & KDD & AuCredit & GeCredit & Heart & ThSurgery & AMPG & \multirow{1}{*}{Contra.}\tabularnewline
\hline 
\hline 
BMM \cite{bouguessa2015practical,do2016outlier} & \textendash{} & 0.97 & 0.93 & 0.87 & 0.94 & 0.62 & 0.67\tabularnewline
ODMAD \cite{do2016outlier,koufakou2008detecting} & \textendash{} & 0.94 & 0.81 & 0.63 & 0.88 & 0.57 & 0.52\tabularnewline
GLM-t \cite{do2016outlier,lud2016iscovering} & \textendash{} & \textendash{} & \textendash{} & 0.72 & \textendash{} & 0.64 & \textendash{}\tabularnewline
Mv.RBM \cite{do2016outlier} & 0.71 & 0.90 & 0.95 & 0.94 & 0.90 & \textbf{1.00} & 0.91\tabularnewline
\hline 
\hline 
$\model$-L2p0.5 & 0.72 & 0.93 & \textbf{0.97} & 0.94 & \textbf{0.97} & \textbf{1.00} & \textbf{0.95}\tabularnewline
$\model$-L2p1 & 0.72 & 0.93 & 0.95 & 0.94 & \textbf{0.97} & \textbf{1.00} & \textbf{0.95}\tabularnewline
$\model$-L2p2 & 0.69 & 0.93 & \textbf{0.97} & 0.94 & \textbf{0.97} & \textbf{1.00} & \textbf{0.95}\tabularnewline
$\model$-L2p$\infty$ & 0.69 & 0.73 & \textbf{0.97} & \textbf{1.00} & \textbf{0.97} & \textbf{1.00} & \textbf{0.95}\tabularnewline
\hline 
\hline 
$\model$-L3p0.5 & \textbf{0.73} & \textbf{0.98} & \textbf{0.97} & 0.94 & \textbf{0.97} & 0.70 & \textbf{0.95}\tabularnewline
$\model$-L3p1 & 0.72 & \textbf{0.98} & \textbf{0.97} & 0.94 & \textbf{0.97} & 0.70 & \textbf{0.95}\tabularnewline
$\model$-L3p2 & 0.71 & \textbf{0.98} & \textbf{0.97} & 0.94 & \textbf{0.97} & 0.70 & \textbf{0.95}\tabularnewline
$\model$-L3p$\infty$ & 0.50 & 0.78 & \textbf{0.97} & 0.94 & \textbf{0.97} & 0.57 & \textbf{0.95}\tabularnewline
\hline 
\end{tabular}
\par\end{centering}
\caption{Anomaly detection F-score on mixed data.\label{tab:F-measure}}
\end{table*}

We use data from \cite{do2016outlier} where the data statistics are
reported in Table~\ref{tab:dataset-prop}. To keep consistent with
previous work \cite{bouguessa2015practical,do2016outlier,koufakou2008detecting,lud2016iscovering},
we report the results using the F-scores. The detection performance
on test data is reported in Table~\ref{tab:F-measure}. Abstraction
works well for $L=2$, where performance is generally better than
the shallow Mv.RBM. However, when one more layer is added, the results
are mixed. This pattern is indeed not new as it resembles what can
be seen across the literature of DBNs for classification tasks \cite{hinton2006rdd,salakhutdinov2009deep}.
One possible conjecture is that at a certain higher level, signals
become too abstract, and that the distribution become too flat to
really distinguish between truly low and high density regions. This
suggests further research on selection of abstraction levels.

\section{Discussion and Conclusion \label{sec:Discussion}}

As an evidence to the argument in Section~\ref{subsec:Separation-of-Abstraction}
about separating the abstraction and detection RBMs, we found that
the sizes of the RBMs that work well on the MNIST do not resemble
those often found in the literature (e.g., see \cite{hinton2006rdd}).
For example, typical numbers of hidden units range from 500 to 1,000
for a good generative model of digits. However, we observe that 10
to 20 units for detection RBMs and 50-100 units for abstraction RBMs
work well in our experiments, regardless of the training size. This
suggests that the number of bits required for data generation is higher
than those required for anomaly detection. This is plausible since
accurate data generation model needs to account for all factors of
variation and a huge number of density modes. On the other hand, anomaly
detection model needs only to identify \emph{low density regions}
regardless of density modes. An alternative explanation is that since
the CD-1 procedure used to train RBMs (see Section~\ref{subsec:Deep-Belief-Nets})
creates \emph{deep energy wells around each data points}, an expressive
model may lead to more false alarms. Thus, the smoothness of the energy
surface may play an important role in anomaly detection. Our $\model$
algorithm offers a consensus among multiple energy surface, and thus
can be considered as a way to mitigate the energy wells issue.

There has been an unexpected connection between the construction procedure
of DBNs and the variational renomarlization groups in physics \cite{mehta2014exact}.
In particular, with layerwise construction, the data is rescaled \textendash{}
the higher layer operates on a coarser data representation. This agrees
with our initial motivation for the MAD hypothesis.

Finally, although not implemented here, the $\model$ lends itself
naturally to detecting anomalies in \emph{multimodal data} with diverse
modal semantics. For example, an image can be equipped with high-level
tags and several visual representations. Each data representation
can be modelled as a Mv.RBM at the right level of abstraction. The
upper RBMs then integrate all information into coherent representations
\cite{tran2016learning}.

\subsection{Conclusion\@.~}

In this paper we have tackled the double challenge of high-dimensions
and mixed-data in anomaly detection. We first proposed the Multilevel
Anomaly Detection (MAD) hypothesis in that a data point is anomalous
with respect to one or more levels of data abstraction. To test the
hypothesis, we introduced $\model$, a procedure to train a sequence
of Deep Belief Networks, each of which provides a ranking of anomalies.
All rankings are then aggregated through a simple $p$-norm trick.
Experiments on both single-type and mixed-type data confirmed that
(a) learning data representation through multilevel abstraction is
a sensible strategy for high-dimensional settings; and (b) $\model$
is a competitive method. There are rooms for improvement however.
First, going very deep has not proved very successful. DBNs have demonstrated
its usefulness in abstraction, but there exist other possibilities
\cite{vincent2010stacked,zhai2016deep}. Finally, the simple $p$-norm
rank aggregation can be replaced by a more sophisticated method for
selecting and building right abstraction levels \cite{zimek2012evaluation}.

\section*{Acknowledgments}

This work is partially supported by the Telstra-Deakin Centre of Excellence
in Big Data and Machine Learning.

\bibliographystyle{plain}

\begin{thebibliography}{10}

\bibitem{aggarwal2001surprising}
Charu~C Aggarwal, Alexander Hinneburg, and Daniel~A Keim.
\newblock On the surprising behavior of distance metrics in high dimensional
  space.
\newblock In {\em International Conference on Database Theory}, pages 420--434.
  Springer, 2001.

\bibitem{aggarwal2015theoretical}
Charu~C Aggarwal and Saket Sathe.
\newblock Theoretical foundations and algorithms for outlier ensembles.
\newblock {\em ACM SIGKDD Explorations Newsletter}, 17(1):24--47, 2015.

\bibitem{ailon2008aggregating}
Nir Ailon, Moses Charikar, and Alantha Newman.
\newblock Aggregating inconsistent information: ranking and clustering.
\newblock {\em Journal of the ACM (JACM)}, 55(5):23, 2008.

\bibitem{ando2015ensemble}
Shin Ando, Theerasak Thanomphongphan, Yoichi Seki, and Einoshin Suzuki.
\newblock Ensemble anomaly detection from multi-resolution trajectory features.
\newblock {\em Data Mining and Knowledge Discovery}, 29(1):39--83, 2015.

\bibitem{angiulli2002fast}
Fabrizio Angiulli and Clara Pizzuti.
\newblock Fast outlier detection in high dimensional spaces.
\newblock In {\em European Conference on Principles of Data Mining and
  Knowledge Discovery}, pages 15--27. Springer, 2002.

\bibitem{barros2015distribution}
Fernando~C Barros, Aris~T Papageorghiou, Cesar~G Victora, Julia~A Noble, Ruyan
  Pang, Jay Iams, Leila~Cheikh Ismail, Robert~L Goldenberg, Ann Lambert,
  Michael~S Kramer, et~al.
\newblock The distribution of clinical phenotypes of preterm birth syndrome:
  implications for prevention.
\newblock {\em JAMA pediatrics}, 169(3):220--229, 2015.

\bibitem{becker2015deep}
John Becker, Timothy~C Havens, Anthony Pinar, and Timothy~J Schulz.
\newblock Deep belief networks for false alarm rejection in forward-looking
  ground-penetrating radar.
\newblock In {\em SPIE Defense+ Security}, pages 94540W--94540W. International
  Society for Optics and Photonics, 2015.

\bibitem{bengio2013representation}
Yoshua Bengio, Aaron Courville, and Pascal Vincent.
\newblock Representation learning: A review and new perspectives.
\newblock {\em IEEE Transactions on Pattern Analysis and Machine Intelligence},
  35(8):1798--1828, 2013.

\bibitem{bouguessa2015practical}
Mohamed Bouguessa.
\newblock A practical outlier detection approach for mixed-attribute data.
\newblock {\em Expert Systems with Applications}, 42(22):8637--8649, 2015.

\bibitem{campos2015evaluation}
Guilherme~O Campos, Arthur Zimek, J{\"o}rg Sander, Ricardo~JGB Campello,
  Barbora Micenkov{\'a}, Erich Schubert, Ira Assent, and Michael~E Houle.
\newblock On the evaluation of unsupervised outlier detection: measures,
  datasets, and an empirical study.
\newblock {\em Data Mining and Knowledge Discovery}, pages 1--37, 2015.

\bibitem{chandola2009anomaly}
Varun Chandola, Arindam Banerjee, and Vipin Kumar.
\newblock {Anomaly detection: A survey}.
\newblock {\em ACM computing surveys (CSUR)}, 41(3):15, 2009.

\bibitem{do2016outlier}
Kien Do, Truyen Tran, Dinh Phung, and Svetha Venkatesh.
\newblock Outlier detection on mixed-type data: An energy-based approach.
\newblock {\em International Conference on Advanced Data Mining and
  Applications (ADMA 2016)}, 2016.

\bibitem{fiore2013network}
Ugo Fiore, Francesco Palmieri, Aniello Castiglione, and Alfredo De~Santis.
\newblock {Network anomaly detection with the restricted Boltzmann machine}.
\newblock {\em Neurocomputing}, 122:13--23, 2013.

\bibitem{gao2014intrusion}
Ni~Gao, Ling Gao, Quanli Gao, and Hai Wang.
\newblock An intrusion detection model based on deep belief networks.
\newblock In {\em Advanced Cloud and Big Data (CBD), 2014 Second International
  Conference on}, pages 247--252. IEEE, 2014.

\bibitem{ghoting2004loaded}
Amol Ghoting, Matthew~Eric Otey, and Srinivasan Parthasarathy.
\newblock Loaded: Link-based outlier and anomaly detection in evolving data
  sets.
\newblock In {\em ICDM}, pages 387--390, 2004.

\bibitem{Hinton02}
G.E. Hinton.
\newblock Training products of experts by minimizing contrastive divergence.
\newblock {\em Neural Computation}, 14:1771--1800, 2002.

\bibitem{hinton2006rdd}
G.E. Hinton and R.R. Salakhutdinov.
\newblock Reducing the dimensionality of data with neural networks.
\newblock {\em Science}, 313(5786):504--507, 2006.

\bibitem{hinton2006fast}
Geoffrey~E Hinton, Simon Osindero, and Yee-Whye Teh.
\newblock A fast learning algorithm for deep belief nets.
\newblock {\em Neural computation}, 18(7):1527--1554, 2006.

\bibitem{kamyshanska2015potential}
Hanna Kamyshanska and Roland Memisevic.
\newblock The potential energy of an autoencoder.
\newblock {\em Pattern Analysis and Machine Intelligence, IEEE Transactions
  on}, 37(6):1261--1273, 2015.

\bibitem{koufakou2008detecting}
Anna Koufakou, Michael Georgiopoulos, and Georgios~C Anagnostopoulos.
\newblock Detecting outliers in high-dimensional datasets with mixed
  attributes.
\newblock In {\em DMIN}, pages 427--433. Citeseer, 2008.

\bibitem{lecun2015deep}
Yann LeCun, Yoshua Bengio, and Geoffrey Hinton.
\newblock Deep learning.
\newblock {\em Nature}, 521(7553):436--444, 2015.

\bibitem{lud2016iscovering}
Yen-Cheng Lu, Feng Chen, Yating Wang, and Chang-Tien Lu.
\newblock Discovering anomalies on mixed-type data using a generalized
  student-t based approach.
\newblock {\em IEEE Transactions on Knowledge and Data Engineering,
  DOI:10.1109/TKDE.2016.2583429}, 2016.

\bibitem{mehta2014exact}
Pankaj Mehta and David~J Schwab.
\newblock An exact mapping between the variational renormalization group and
  deep learning.
\newblock {\em arXiv preprint arXiv:1410.3831}, 2014.

\bibitem{salakhutdinov2009deep}
R.~Salakhutdinov and G.~Hinton.
\newblock {Deep Boltzmann Machines}.
\newblock In {\em Proceedings of 20th AISTATS}, volume~5, pages 448--455, 2009.

\bibitem{sun2014automated}
Jianwen Sun, Reto Wyss, Alexander Steinecker, and Philipp Glocker.
\newblock Automated fault detection using deep belief networks for the quality
  inspection of electromotors.
\newblock {\em tm-Technisches Messen}, 81(5):255--263, 2014.

\bibitem{tagawa2014structured}
Takaaki Tagawa, Yukihiro Tadokoro, and Takehisa Yairi.
\newblock Structured denoising autoencoder for fault detection and analysis.
\newblock In {\em ACML}, 2014.

\bibitem{Truyen:2011b}
T.~Tran, D.Q. Phung, and S.~Venkatesh.
\newblock {Mixed-variate restricted Boltzmann machines}.
\newblock In {\em Proc. of 3rd Asian Conference on Machine Learning (ACML)},
  Taoyuan, Taiwan, 2011.

\bibitem{tran2016preterm}
Truyen Tran, Wei Luo, Dinh Phung, Jonathan Morris, Kristen Rickard, and Svetha
  Venkatesh.
\newblock Preterm birth prediction: Deriving stable and interpretable rules
  from high dimensional data.
\newblock {\em Conference on Machine Learning in Healthcare, LA, USA}, 2016.

\bibitem{tran2016learning}
Truyen Tran, Dinh Phung, and Svetha Venkatesh.
\newblock Learning deep representation of multityped objects and tasks.
\newblock {\em arXiv preprint arXiv:1603.01359}, 2016.

\bibitem{vincent2010stacked}
P.~Vincent, H.~Larochelle, I.~Lajoie, Y.~Bengio, and P.A. Manzagol.
\newblock Stacked denoising autoencoders: Learning useful representations in a
  deep network with a local denoising criterion.
\newblock {\em The Journal of Machine Learning Research}, pages 3371--3408,
  2010.

\bibitem{vovsha2014predicting}
Ilia Vovsha, Ashwath Rajan, Ansaf Salleb-Aouissi, Anita Raja, Axinia Radeva,
  Hatim Diab, Ashish Tomar, and Ronald Wapner.
\newblock Predicting preterm birth is not elusive: Machine learning paves the
  way to individual wellness.
\newblock In {\em 2014 AAAI Spring Symposium Series}, 2014.

\bibitem{wang2016deep}
Yao Wang, Wan-dong Cai, and Peng-cheng Wei.
\newblock {A deep learning approach for detecting malicious JavaScript code}.
\newblock {\em Security and Communication Networks}, 2016.

\bibitem{zhai2016deep}
Shuangfei Zhai, Yu~Cheng, Weining Lu, and Zhongfei Zhang.
\newblock Deep structured energy based models for anomaly detection.
\newblock {\em arXiv preprint arXiv:1605.07717}, 2016.

\bibitem{zhang2010effective}
Ke~Zhang and Huidong Jin.
\newblock An effective pattern based outlier detection approach for mixed
  attribute data.
\newblock In {\em Australasian Joint Conference on Artificial Intelligence},
  pages 122--131. Springer, 2010.

\bibitem{zimek2012survey}
Arthur Zimek, Erich Schubert, and Hans-Peter Kriegel.
\newblock A survey on unsupervised outlier detection in high-dimensional
  numerical data.
\newblock {\em Statistical Analysis and Data Mining}, 5(5):363--387, 2012.

\bibitem{zimek2012evaluation}
Erich Schubert Remigius Wojdanowski~Arthur Zimek and Hans-Peter Kriegel.
\newblock On evaluation of outlier rankings and outlier scores.
\newblock 2012.

\end{thebibliography}

\end{document}